\documentclass[runningheads]{llncs}

\usepackage[frozencache,cachedir=.]{minted}
\usepackage{algorithm}
\usepackage{algpseudocode}
\usepackage{subfigure}
\usepackage{graphicx}

\usepackage{cite}
\begin{document}
\title{Automating UAV Flight Readiness Approval using Goal-Directed Answer Set Programming}
\titlerunning{Flight Readiness Approval using sCASP}
%
\author{Sarat Chandra Varanasi\inst{1,2}
\and
Baoluo Meng\inst{1} \and
Christopher Alexander\inst{1}\and
Szabolcs Borgyos\inst{1} \and
Brendan Hall\inst{1}}
\authorrunning{SC Varanasi et al.}
%
\institute{General Electric Research, Niskayuna, NY, USA \\
\email{\{saratchandra.varanasi, baoluo.meng, christopher.alexander, szabolcs.borgyos, brendan.hall\}@ge.com}\\
\and
The University of Texas at Dallas, Richardson, TX, USA\\
\email{sxv153030@utdallas.edu}}
\maketitle              
\begin{abstract}
We present a novel application of Goal-Directed Answer Set Programming that digitizes the model aircraft operator’s compliance verification against the Academy of Model Aircrafts (AMA) safety code.
The AMA safety code regulates how AMA flyers operate Unmanned Aerial Vehicles (UAVs) for limited recreational purposes.
Flying drones and their operators are subject to various rules before and after the operation of the aircraft to ensure safe flights. 
In this paper, we leverage Answer Set Programming to encode the AMA safety code and automate compliance checks.
To check compliance, we use the s(CASP) which is a goal-directed ASP engine.
By using s(CASP) the operators can easily check for violations and obtain a justification tree explaining the cause of the violations in human-readable natural language. 
Further, we implement an algorithm to help the operators obtain the minimal set of conditions that need to be satisfied in order to pass the compliance check. 
We develop a front end questionnaire interface that accepts various conditions and use the backend s(CASP) engine to evaluate whether the conditions adhere to the regulations.
We also leverage s(CASP) implemented in SWI-Prolog, where SWI-Prolog exposes the reasoning capabilities of s(CASP) as a REST service.
 To the best of our knowledge, this is the first application of ASP in the AMA and Avionics Compliance and Certification space.

\end{abstract}
\section{Introduction}
We present a novel application of Answer Set Programming (ASP) that hastens the Academy of Model Aircraft (AMA) Flight Safety Compliance checking process. The AMA Safety Code Compliance rules are written as a set of English language rules describing the situations under which violations can occur. We capture the different sets of conditions that could lead to violations and translate the violations into ASP rules. Exceptional conditions in the rules can be directly mapped to ASP's support for negation-by-failure. We have developed an application in the form of a questionnaire detailing various conditions which require a compliance check. We have used the AMA Safety Code rules, which are simple enough to be coded as propositional answer set programs. That is, the entire questionnaire is a series of yes-or-no questions that can be completed in less than five minutes by an experienced aircraft operator applying for certification.  The s(CASP) system is capable of printing a justification tree for any identified violations which is presented in plain English. This helps the user understand the cause of the violation without having to understand ASP rules. We also developed a feature in our app that recommends the minimal set of conditions that need to be satisfied in order to achieve compliance given the current state. All such explanations are rendered in English and completely abstract the existence of a logical backend reasoner. The effect of this is to provide a robust tool to make sense of the AMA rules and address the slow compliance verification process.  The s(CASP) system provides a goal-directed implementation of ASP and has many advantages over traditional Logic Programming systems and Answer Set Programming implementations \cite{gupta2017arcade, arias2018constraint}. Traditional logic programming systems cannot handle circular rules (with or without negations) present in a knowledge base. ASP can handler circular rules (containing negations) using stable model semantics and enables the capture of complex reasoning \cite{gelfond2014knowledge, erdem2016applications, arias2018constraint}. Further, goal-directed ASP systems such as s(CASP), only search the part of the knowledge base relevant to a query, thus making the complex reasoning scalable \cite{arias2018constraint}. 
\par
Our main contribution is the following, we have translated general AMA guidelines into ASP and thus captured the knowledge present in the guidelines for reasoning. We have used the state-of-the-art goal-directed s(CASP) system to produce justification trees for guideline violations. We have also developed an end-to-end web application that first asks the self-certifier (user of the application) a series of questions to self-check any violations. If there are any violations, a minimal set of conditions that need to be changed is presented to the user in English. The application makes the entire process of self-compliance transparent without knowledge of the complex reasoning involved in s(CASP).To the best of our knowledge, this is the first application of ASP in aviation rules compliance checking. 
\section{Background}

\subsection{Answer Set Programming}
Answer Set Programming (ASP) is a declarative knowledge-representation and reasoning paradigm with several applications in industrial planning and optimization \cite{erdem2016applications, gelfond2014knowledge}. ASP enables one to perform commonsense reasoning and abductive reasoning in particular. 
An ASP program comprises of rules of the form $p \leftarrow q_{1}, q_{2}, ... q_{m},  not \ r_{1},  ..., not \ r_{n}$, where $m \geq 0, n \geq 0$, and $p, q_{1}, q_{2}, ... q_{m}, r_{1},..., r_{n}$ are proposition literals. The operator \textit{not} represents the negation-by-failure operator. The above rule form means that $p$ is true, if $q_1, q_2, .., q_m$ are true are $r_1,  ..., r_n$ do not have a proof. The proposition on the left hand side of the left arrow constitute the \textit{head} of a rule and the propositions on the right-hand side of the arrow (including the propositions qualified by $not$) constitute the \textit{body} of the same rule. The stable models (or answer sets) of an ASP program are the sets of satisfiable propositions in the given program. 
ASP also allows definition of predicates with variables. In such a case, ASP solvers propositionalize the program by substituting every possible value for each variable, a process termed \textit{grounding}. For example, if the knowledge base has the rule: $p(X, Y) \leftarrow q(X), not \ r(Z)$, then ASP-solvers such as Clingo \cite{gebser2018potsdam} ground the program into propositions by substituting every possible value for variables $\{X, Y, Z\}$. 
ASP allows one to define complicated rules, particularly set of rules that allow circular definitions through negations. For example, the rules $p \leftarrow not \ q $ and $q \leftarrow not \ p$ are circular rules through negation. In such a case, ASP produces two models, one model in which $p$ is true and another in which  $q$ is true. This makes ASP a \textit{non-monotonic} logic.  When the head of a rule is empty, then it represents a head-less rule or a constraint. A rule such as $f\!alse \leftarrow q_1, q_2, .., q_m, not \ r_1, not \ r_2, ..., not \ r_n$ means that $q_1, q_2, ..., q_m, not \ r_1, not \ r_2, ..., not \ r_n$ cannot all be simultaneously true. Abductive reasoning can be performed using even loops through negation. Even loops through negation represent circular rules where there are an even number of intervening negations from a rule head to itself. For example, $p \leftarrow not \ q$ and $q \leftarrow not \ p$ is an even loop through negation for both $p$ and $q$. The even loop can also mean that, one is \textit{assuming} $p$ to be either true or false, that is, they are \textit{abducing} $p$ \cite{arias2018constraint}.
\subsection{Goal-Directed Answer Set Programming}
\begin{figure}[ht]
 \centering
 \includegraphics[scale=0.8]{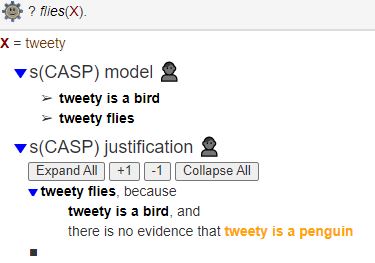}
 \caption{An English justification tree that tweety flies}
 \label{fig:sample_tree}
\end{figure}

There are alternative approaches to Answer Set Solving that do not perform grounding to find stable models. One such approach is Goal-Directed ASP\cite{gde2021}. Goal-directed ASP assumes that a query $?- p(X_1, X_2, ..., X_n)$ is provided to the ASP solver. The goal-directed search finds the predicates that support the submitted query and finds bindings for $X_1, X_2, .., X_n$. At an interactive level, this looks similar to Logic Programming, however, the underlying search uses the co-SLDNF resolution algorithm to find stable models\cite{min2009predicate, arias2018constraint}. A salient feature of goal-directed ASP is that, the co-SLDNF algorithm only searches for rules that are relevant to finding the support for the given query. The state-of-the-art implementation of goal-directed ASP is the s(CASP) system. When given a query $?- q$, s(CASP) returns a \textit{partial stable model} of $q$ for the given ASP program. Similar to Prolog, the user can press a semi-colon to print more models. If a proposition $p$ is not present in any of the partial models, then it is not present in any complete model of the program. If indeed, $p$ is present in some partial model, then there exists a complete model where both $p$ and $q$ are true. More details about partial models can be found elsewhere \cite{arias2018constraint}. 
Along with the partial stable model, s(CASP) also prints a justification tree to the user conveying how the proof for the given query was performed. The justifications of ASP queries have been used widely in several application areas such as Explainable AI, Legal Reasoning and Natural Language Understanding \cite{arias2021modeling, basu2020aqua, basu2020square, basu2021knowledge}. Further, the justifications can also be rendered in human understandable natural language \cite{arias2020justification}. For example, consider the below ASP program with the predicates mapped to their English descriptions. An example justification tree (s(CASP) justification tree) and partial stable model (s(CASP) model) for the program below is shown in Figure \ref{fig:sample_tree}. The \texttt{'@(X)'} notation is used by s(CASP) to substitute the value that the variable \texttt{X} unifies with, at the time of program execution. Even ASP programs that only contain propositions can be mapped to their English translations in the same way as shown. 
\begin{minted}{prolog}
flies(X) :- bird(X), not penguin(X).
bird(tweety).
#pred flies(X)::'@(X) flies'.
#pred bird(X)::'@(X) is a bird'.
#pred penguin(X): '@(X) is a penguin'.
\end{minted}

Similar to ASP, abduction can be easily performed in s(CASP). To abduce a literal $p$, the directive $\#abducible \ p$ can be declared in the ASP program. This directive is a short-hand for the even loop through negation on $p$, where another literal $p'$ not part of the original program is used as part of the even loop \cite{arias2018constraint}. This mechanism enables the use of ``hidden'' propositions not visible to the author of the ASP program. Every abducible has its own unique hidden proposition generated by the ASP solver to avoid inconsistencies.  

\subsection{Modern SWISH Interface API for s(CASP)}
The s(CASP) system is available on the modern SWISH Interface for SWI-Prolog \cite{wielemaker2019using, wielemaker2021scasp}. SWISH allows a user to create interactive Logic Program Notebooks for s(CASP) answer set programs. The s(CASP) module can be enabled by using the \mintinline{prolog}{:- use_module(library(scasp)).} Along with the online interface, SWISH also provides REST\footnote{REST APIs are standard HTTP apis, more details about the API can be found here\url{https://www.swi-prolog.org/pldoc/doc\_for?object=section(packages/pengines.html)}} (Representational State Transfer) APIs for external clients to run ASP programs. The AMA Safety Code application in this paper uses the SWISH API for s(CASP).  

We next explain the Flight Compliance and Certification process used to ensure compliance by recreational UAV operators with AMA rules, followed by their translation in ASP. We also show snippets of the user interface involved in the certification process. We then describe a simple algorithm to compute minimal explanation required to achieve compliance, in cases where a violation exists. Finally we mention related work and conclude by providing directions for future work. 

\section{The AMA Safety Code for Aircraft Operators}
The Academy of Model Aeronautics (AMA) defines a set of rules and regulations to be followed by the operator(s) of model aircraft during their flight. We have adopted the rules defined in the AMA Safety code\cite{ama_rules}. The AMA Safety Code has conditions pertaining to general operation of aircrafts along with regulations pertaining to radio control and free flight. The rules are written in plain english. Each rule either prescribes a certain set of conditions desirable for flight safety or proscribes a set of conditions that violate safety requirements. We encode each rule as a violation rule. A few AMA safety code rules are shown below.
\begin{itemize}
    \item I will not fly a model aircraft in a careless or reckless  manner 
    \item I will not interfere with and will yield the right of way to all human-carrying  aircraft using AMA’s See and Avoid Guidance and a spotter when appropriate
    \item I will not operate any model aircraft while I am under the influence of alcohol or any drug that could adversely affect my ability to safely control the model.
\end{itemize}

For example, the AMA rule 3 states that \textit{ I will not operate my model aircraft while under the influence of alcohol or while using any drug which could adversely affect my ability to safely control the model}. This potential violation is capture by a rule in ASP as: \mintinline{prolog}{violation_3 :- alcohol_or_drug_influence}. 

Many of the rules are written in terms of \textit{defaults} and \textit{exceptions}. In such case, the rules will be of the form, \mintinline{prolog}{some_violation :- default, not exception}. 

\begin{figure}[ht]
 \centering
 \includegraphics[scale=0.3]{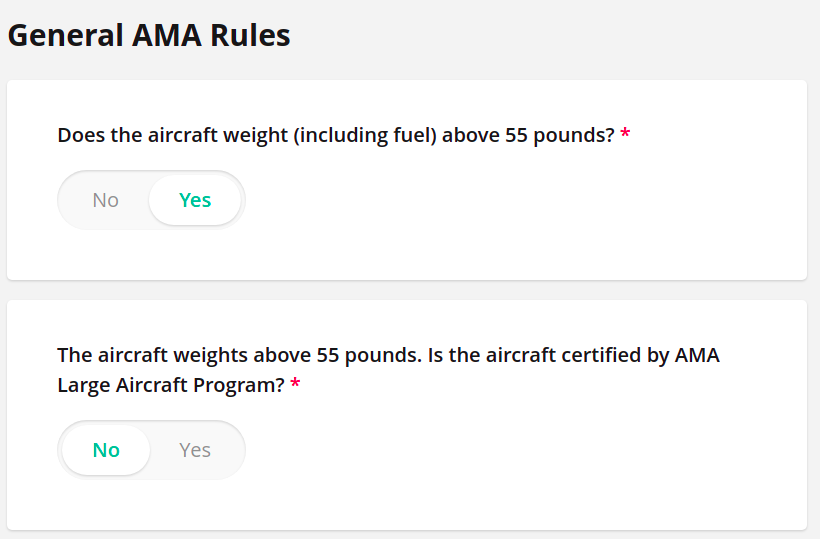}
 \caption{Questionnaire corresponding to aircraft weight}
 \label{fig:sample_question}
\end{figure}

For instance, first clause of rule 7 states that: \textit{I will only fly models weighing more than 55
pounds, including fuel, if certified through AMA’s 
Large Model Airplane Program}.  The sub-clause before \textit{if} represents a default and the sub-clause after \textit{if} represents the exception to the default. This violation is encoded as: \\ \mintinline{prolog}{violation_7 :- aircraft_above_55, not certified_by_ama_large_program}.

\begin{figure}[!tbp]
 \centering
 \includegraphics[scale=0.4]{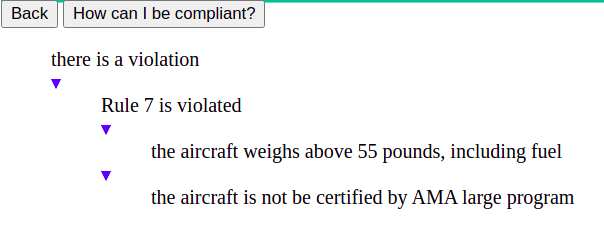}
 \caption{Violation:aircraft weighs above 55 pounds, not certified by AMA Large aircraft program}
 \label{fig:sample_violation}
\end{figure}

The AMA flight compliance rules are therefore translated into ASP.  
On the front-end of the compliance application, the questionnaire elicits answers from the user by asking the following sequence of questions. Each of the questions maps to the literals in the body of a potential violation rule.  Shown in figure \ref{fig:sample_question}.  The proof tree returned by s(CASP) is rendered graphically on the front-end \ref{fig:sample_violation}. From the viewpoint of the end-user, the ASP translation and proof tree are completely transparent. Showing the proof graphically helps the operator easily understand the violations and also to figure out how to achieve compliance.

\begin{figure}
    \centering
    \includegraphics[scale=0.15]{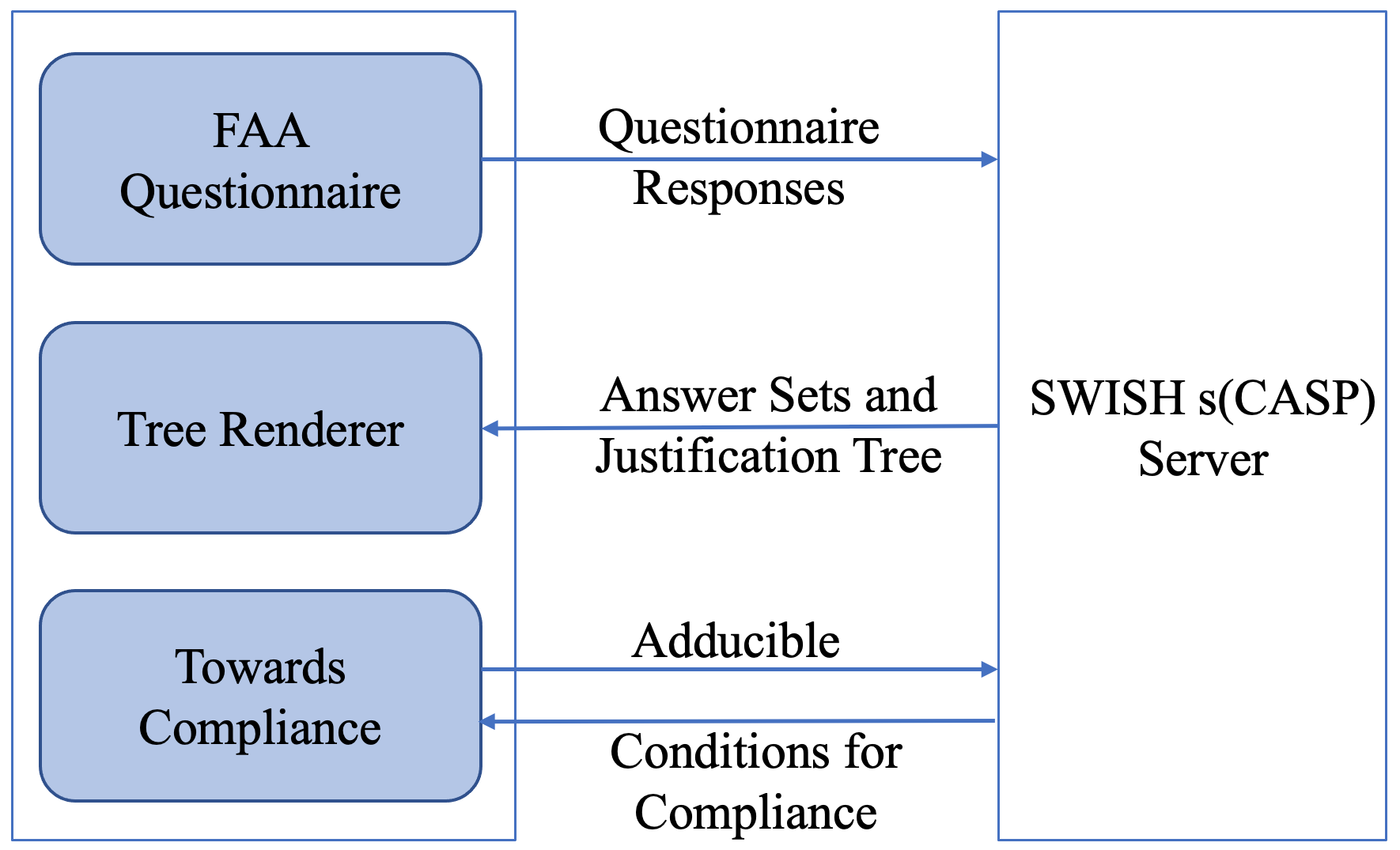}
    \caption{Architecture of the Flight Readiness Approval Application}
    \label{fig:arch}
\end{figure}

The architecture of the web application and its backend interface to SWISH is shown in Figure \ref{fig:arch}. The computation of minimal explanation is discussed in the next section. The app's source itself is accessible from github \cite{github}. The entire AMA General Rules are shown in the listing below. Our experiments were run on a Quad core Intel(R) Core(TM) i7-10510U CPU @1.80Ghz with 8-GB RAM. The average time taken for these rules to be executed, returned the SWISH server is ~86 milliseconds. The time taken to compute minimal explanations towards compliance, in the presences of a rule violation is also similar. 
We have captured only the General AMA rules which are a set of 10 possible violations. They are written into 15 ASP rules. However more rules can be captured from the AMA guidelines. The number of English rules can be quite large and their translation into ASP much larger in practice \cite{chen2016physician}, and yet be handled efficiently due to the goal-directed execution of s(CASP).  
\begin{minted}[frame=single,fontsize=\scriptsize]{prolog}
% Rule 1:  I will not fly a model aircraft in a careless  or reckless  manner 
  violation_1 :-  careless_reckless. 
  careless_reckless :- has_prior_history_of_violation.
% Rule 2: I will not interfere with and will yield the right of way to all  human-carrying  
% aircraft using AMA’s  See and  Avoid  Guidance and a spotter when appropriate
  violation_2 :- human_carrying_aircraft, not yield_right_of_way. 
  violation_2 :- human_carrying_aircraft, yield_right_of_way, not used_spotter.  
% Rule 3: I will not operate any model aircraft while I am under the influence of alcohol or 
% any drug that could adversely affect my ability to safely control the model.
  violation_3 :- alcohol_drug_influence.
% Rule 4: I will avoid flying directly over unprotected people, moving vehicles, and 
% occupied structures.
  violation_4 :- directly_over_people_vehicles_structures.
% Rule 5: I will fly Free Flight (FF) and Control Line (CL) models in compliance with AMA’s 
% safety programming.
    violation_5 :- ff_cl, not ama_safety_programming.
% Rule 6: I will maintain visual contact of an RC model aircraft without enhancement other 
% than corrective lenses prescribed to me. When using an advanced flight system,  such as an 
% autopilot, or flying First-Person View (FPV), I will comply with AMA’s Advanced Flight 
% System programming.
  violation_6 :- visual_contact_using_enhancement
  violation_6 :- advanced_system_or_fpv, ama_advanced_flight_system_programming.
% Rule 7: I will only fly models weighing more than 55 pounds, including fuel, if certified  
% through AMA’s Large Model Airplane Program
  violation_7 :-  aircraft_weighs_above_55_pounds, not certified_by_ama_large_program.
% Rule 8:I will only fly a turbine-powered model aircraft  in compliance with AMA’s Gas 
% Turbine Program.
  violation_8 :- turbine_model, not ama_gas_turbine_program.
% Rule 9: I will not fly a powered model outdoors closer than 25 feet to any  individual, 
% except for myself or my helper(s) located at the flightline, unless I am taking off and 
% landing, or as otherwise provided in AMA’s Competition Regulation
  violation_9 :- closer_than_25_ft, not landing_takeoff.
  violation_9 :-  closer_than_25_ft, landing_takeoff, not ama_competition_regulation.
% Rule 10: I will use an established safety line to separate all model aircraft operations 
% from spectators and bystanders.  
  violation_10 :- uses_established_safety_line.
\end{minted}

\section{Computing Minimal Explanation to Compliance}
 
 Based on the AMA questionnaire, when a violation is identified by s(CASP), the user would want to know what is the minimal set of conditions under which the violation can be mitigated to achieve compliance. This is done by computing what is required to satisfy the rule and compare it against the corresponding violation. A simple algorithm is presented. The minimal set of conditions are the sets of propositions whose truth value must be changed from the AMA questionnaire. Before going into notations, a few assumptions are listed:
 \begin{enumerate}
     \item The compliance rules are encoded as a propositional answer set program. That is, there are no variables involved in the rules
     \item The rules are a set of violation definitions, that is, every violation from the compliance rules is numbered by the rule number found in the compliance document. That is, if $rule \ n$ states a violation, then the ASP program would have a rule with head $violation_n$. 
     \item For a given answer set program $\Pi$, $Models(\Pi)$ denotes the set of all answer sets of $\Pi$
     \item For a given answer set program $\Pi$, $Models_{q}(\Pi)$ denotes the set of all partial models for query $?- \  q$.
     \item A response to a question in the questionnaire corresponds to an input literal fact. For example, if a user selects ``Yes" for \textit{Was the operator under influence of alcohol or drugs?}, then the input literal fact \texttt{alcohol\_drug\_influence} is added to $\Pi$.  If the user selects ``No", then the literal fact is not added.
     \item The literals are also referred to as conditions in the discussion below.
 \end{enumerate}

\begin{figure}
    \centering
    \includegraphics[scale=0.4]{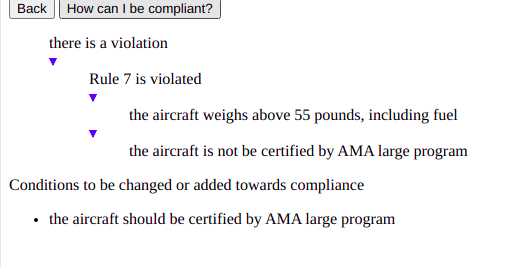}
    \caption{Minimal conditions to be satisfied to comply with Rule 7}
    \label{fig:sample_compliance}
\end{figure}

 Let $\mathcal{C}$ denote the set of conditions from a questionnaire. Conditions are the facts collected by the questionnaire and can be treated as propositions.  Let $\Pi$ denote the ASP program encoding compliance rules. Let $\mathcal{C}_{violate}$ denote the set of conditions that enable some violation $violation_i$. That is, $violation_i \in PartialModel$ such that $PartialModel \in Models(\Pi \cup \mathcal{C}_{violate})$. A simple approach to find the conditions towards compliance is to query $?- not \ violate_i$ to s(CASP) and gather its partial stable models. However, this will not guarantee that the conditions will be found. For example, if $violate_i \leftarrow not \ c$, then the dual for $violate_i $ will be $not \ violate_i \leftarrow c$. However, directly querying for $?- not \ violate_i$ will cause a failure as $c$ needs to be populated from the questionnaire. To address this, $c$ can be treated as an abducible. Then, s(CASP) will abduce one world in which $c$ would be true, and hence find support for $?- not \ violate_i$. Therefore, every condition $c \in C$ would be treated as an abducible and then $?- not \ violate_i$ can be queried. The partial model $\mathcal{C}_{\neg violate}$ whose literals have the least difference to the literals of $\mathcal{C}_{violate}$ is computed. The truth values of literals from $\mathcal{C}_{\neg violate}$ represent the minimal set of conditions to achieve compliance, while retaining as may decisions selected by the user. The minimal set of conditions to be compliance for violating rule 7 (from Figure \ref{fig:sample_violation}) is shown in Figure \ref{fig:sample_compliance}
\begin{algorithm}
\caption{Compute minimal explanation to enable compliance of $violation_i$}
\begin{algorithmic}
\Require $violation_i \in PartialModel \land Partial Model \in Models(\Pi)$
\Ensure Compute $\mathcal{C}_{\neg violation_i}$ where $not \ violation_{i} \in Models(\Pi)$ 
\ForAll {$literal \ c \in \mathcal{C}$}
  \State $\Pi \gets \Pi \cup \{\#abducible \ c. \}$
\EndFor
\State $PartialModels \gets Models_{(not \ violation_i)}(\Pi)$
\State Find $model \in PartialModels$ \\ 
 ~~~~~~~~~~~such that $model \cap \mathcal{C}_{violate} \neq \emptyset \land | model \setminus {C}_{violate} | $ is smallest  
\State \Return  $model$
\end{algorithmic}
\end{algorithm}
\section{Related Work}
  Several works exist in literature towards transforming compliance rules into Answer Set Programs  \cite{d2011verifying, vos2019odrl}. The work of Marina De Vos et. al \cite{vos2019odrl} translates ODRL rules into ASP programs whereas work of Davide D'Aprile et al. \cite{d2011verifying} uses Action Languages to model business processes in Answer Set Programming. Action Languages allows one to model \textit{change}, and temporal aspects associated with \textit{cause-effect reasoning}. Action Languages are easily translatable into ASP\cite{gelfond2014knowledge}. Seminal work in the area of goal-directed ASP is the Physician Advisory system (CHeF system) that models ASP rules for the diagnosis and treatment of congestive heart failure \cite{chen2016physician} based on expert rules from the American College of Cardiology Foundation and American Heart Association.  CHeF also proposes several knowledge patterns that are modeled as ASP rules to perform congestive heart failure diagnosis. Textual Rulelog is a major declarative logic programming language that can model compliance rules and provide explanations. It has been applied to model Regulation W of US Federal Reserve Laws among many more applications \cite{grosof2014automated}. Textual Rulelog is based on the well-founded semantics whereas ASP is based on the stable model semantics of logic programs. Work in goal-directed ASP has resulted in the unification of stable model and well-founded semantics. When the knowledge base contains loops through negation, then stable model semantics splits the loops through negation into different worlds and can still perform reasoning. However, well-founded semantics treats the truth-value of propositional literals as \textit{unknown} when they are involved in a loop through negation \cite{salazar2016proof}.    

\section{Conclusion and Future Work}
 We have developed a novel flight safety application that uses state-of-the-art goal-directed s(CASP) system.
 The justifications for non-compliance are shown in human understandable natural language. This makes the end user totally oblivious of the intricacies involved in the s(CASP) backend. We also provided an algorithm that computes the minimal set of conditions that need to be changed in order to achieve compliance. This feature provides the user a better understanding of the AMA rules to regulate self-compliance.  The explanations are also rendered in natural language. We plan to incorporate more rules that are part of the  certification process. 
 
The AMA rules previously described in this paper apply to recreational UAV. The application can also be tailored for use in commercial operations compliance verification which is more strictly enforced. The FAA’s Part 107 (P107) federal legislation established guidelines for all small unmanned aircraft system (sUAS) commercial operations to ensure safe and efficient aviation operations across the U.S. The legislation dictates a restriction on operations using a binary but sometimes subjective decision tree structure which has been enforced through certification actions, civil, and criminal penalties. Many P107 rules are inherently subjective without a standard measure of compliance. For example, 107.37(b) states “No person may operate a small unmanned aircraft so close to another aircraft as to create a collision hazard.“ where “so close” and “collision hazard” are not well defined and there is no guidance provided on how the operator should asses the requirement. sUAS operators must navigate the nuances of numerous rules strategically and also tactically during flight. All commercial sUAS operators are required to obtain a P107 license certifying the pilot is able to self-regulate and defend their operational intent in the event of a future complaint investigation or FAA enforcement action. It is largely the responsibility of the operator to fully understand their operating environment (both physical and technical) and prove P107 rules are applied appropriately. 

 Commercial operators intending to fly outside of the P107 restrictions must submit a waiver describing intent and infrastructure capabilities to the FAA for review, with an expected response time of ninety days. The FAA must review the waiver and consider the operational landscape as well as the technicalities of the request to determine if any configuration could support the flight. This inefficient and time-consuming process could lead to delays in flight operations.
 \cite{mavin2009easy}. Automating P107 is part of our future work. 

\bibliographystyle{splncs04}
\bibliography{references}




\end{document}